\documentclass[10pt,twocolumn,letterpaper]{article}

\usepackage{cvpr}
\usepackage{times}
\usepackage{epsfig}
\usepackage{graphicx}
\usepackage{amsmath}
\usepackage{amssymb}
\usepackage{multirow}
\usepackage[table]{xcolor}
\usepackage{booktabs}
\usepackage{color}
\usepackage{subfigure}
\usepackage{booktabs}

\usepackage{algorithm}
\usepackage{algorithmic}

\definecolor{green}{RGB}{0,128,0}
\definecolor{brown}{RGB}{128,0,41}

\newcommand{\tabincell}[2]{
\begin{tabular}{@{}#1@{}}#2\end{tabular}
}

\usepackage[pagebackref=true,breaklinks=true,letterpaper=true,colorlinks,bookmarks=false]{hyperref}

\cvprfinalcopy 

\def\cvprPaperID{5163} 

\ifcvprfinal\fi
\begin{document}

\title{Partial Order Pruning: for Best Speed/Accuracy Trade-off in \\ Neural Architecture Search}

\author{Xin Li$^{1}$ \quad Yiming Zhou$^{1,3}$ \quad Zheng Pan$^{1}$ \quad Jiashi Feng$^{2}$ \\
\small $^1$UISEE Technology Inc. \\ 
\small $^2$ Department of ECE, National University of Singapore \\
\small  $^3$ National Key Lab. of Communications, UESTC \\
\small \{xin.li,\ yiming.zhou,\ zheng.pan\}@uisee.com \quad elefjia@nus.edu.sg
}

\maketitle

\begin{abstract}
Achieving good speed and accuracy trade-off on a target platform is very important in deploying deep neural networks in real world scenarios. However, most existing automatic architecture search approaches only concentrate on high performance. In this work, we propose an algorithm that can offer better speed/accuracy trade-off of searched networks, which is termed ``Partial Order Pruning''. It prunes the architecture search space with a partial order assumption to automatically search for the architectures with the best speed and accuracy trade-off. Our algorithm explicitly takes profile information about the inference speed on the target platform into consideration. With the proposed algorithm, we present several Dongfeng (DF) networks that provide high accuracy and fast inference speed on various application GPU platforms. By further searching decoder architectures, our DF-Seg real-time segmentation networks yield state-of-the-art speed/accuracy trade-off on both the {target embedded device} and the high-end GPU.
\vspace{-0.5cm}
\end{abstract}

\section{Introduction}
Deploying deep convolutional neural networks (CNNs) on real-world embedded devices is attracting increasing research interest. 
Different from high-end GPUs, these devices usually offer rather limited computation capacity, leading to low efficiency when deploying popular high accuracy CNN models~\cite{he2016deep,chen2018deeplab}
on them. 
Despite the considerable efforts on accelerating inference of CNNs such as
pruning~\cite{guo2016dynamic}, quantization~\cite{wu2016quantized} and factorization~\cite{liu2015sparse},  fast inference speed\footnote{Inference speed is measured by inference latency, which is defined as inference time with batch size 1 in a CNN.} is usually achieved 
at the cost of degraded performance~\cite{paszke2016enet,krizhevsky2012imagenet}.
In this paper, we address such a practical problem: Given a target platform, what is the best speed/accuracy trade-off boundary curve by varying CNN architecture? Or
more specifically, we aim to answer two questions:
1) Given the maximum acceptable latency, what is the best accuracy one can get? 
2) To meet certain accuracy requirements, what is the lowest inference latency one can expect?

\begin{figure}[!t]
	\centering
	\includegraphics[width=0.85\linewidth]{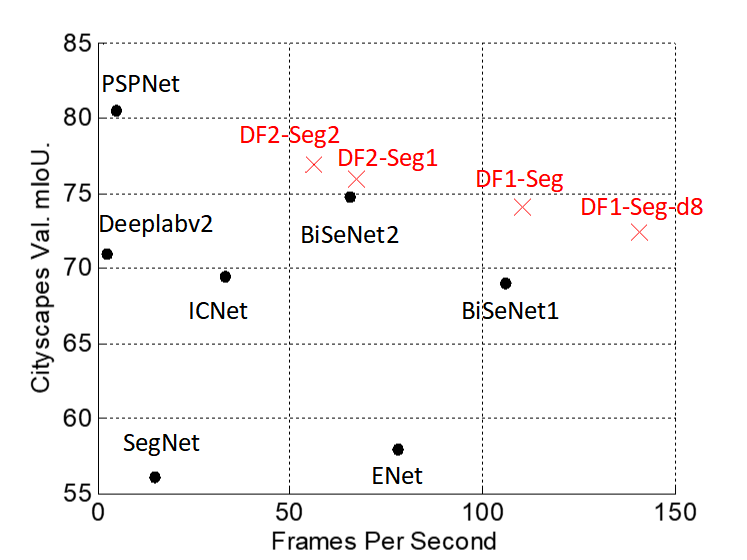}
	\caption{\small Speed (frames per second) and segmentation accuracy ($\text{mIoU}_\text{class}$) comparison  on the Cityscapes~\cite{cordts2016cityscapes} validataion set. DF-Seg networks yield state-of-the-art speed/accuracy trade-off. The compared models include  PSPNet~\cite{zhao2017pyramid}, Deeplabv2~\cite{chen2018deeplab}, ENet~\cite{paszke2016enet}, SegNet~\cite{badrinarayanan2017segnet}, ICNet~\cite{zhao2018icnet}, BiSeNet~\cite{yu2018bisenet} and our DF-Seg networks.}
	\label{fig:segmentation_comparison}
	\vspace{-6mm}
\end{figure}

Some existing works manually design high accuracy network architectures~\cite{zhao2018icnet,yu2018bisenet,howard2017mobilenets,sandler2018inverted}. They usually adopt an indirect metric, \emph{i.e.} FLOP, to estimate the network complexity, but the FLOP count does not truly reveal the actual inference speed. 
For example, for a $3\times 3$ convolution on Nvidia GPUs which is highly optimized in terms of both hardware and software design~\cite{lavin2016fast}, one can assume it is $9$ times slower than a $1\times 1$ convolution on GPUs since it has $9$ times more FLOPs, which is not true actually.
Besides, another important factor that affects the inference speed, the memory access, is not covered by measuring FLOPs.
Considering the diversities of hardware and software, it is almost impossible to find one single architecture that is optimal for all the platforms.

Some other works attempt to automatically search for the optimal network architecture~\cite{zoph2018learning,real2018regularized,darts}, but they also rely on FLOPs to estimate the network complexity and do not take into account the discrepancy of this metric with the actual inference speed and also the target platforms. 
{Despite a few works~\cite{dong2018dpp,cai2018proxylessnas,fbnet} consider the actual inference speed on target platforms, they search the architecture in each individual building block and keep fixed the overall architecture, \emph{i.e.} depth and width.}

In this paper, we develop an efficient architecture search algorithm that automatically selects the networks that offer better speed/accuracy trade-off on a target platform.  
The proposed algorithm is termed ``Partial Order Pruning'', with which some candidates that fail to give better speed/accuracy trade-off are filtered out at early stages of the architecture searching process based on a partial order assumption (see Section~\ref{subsecpoa} for details).
For example, a wider network cannot be more efficient than a narrower one with the same depth, thus accordingly some wider ones are discarded. 
By pruning the search space in this way, our algorithm is forced to concentrate on those architectures that are more likely to lift the boundary of speed/accuracy trade-off.

The proposed ``Partial Order Pruning" algorithm differs from previous neural architecture search algorithms in three aspects. Firstly, 
it explicitly takes platform characteristics into consideration. Secondly, it balances the width and depth of the overall architecture, instead of searching for complicated building blocks. Thirdly, it employs a partial order assumption and a cutting plane algorithm to accelerate searching, instead of using reinforcement learning, evolutionary algorithms or gradient-based algorithms. 

With the proposed algorithm, we are able to obtain a set of networks that provide better accuracy and faster inference speed on a target platform, which we call Dongfeng (DF) networks. 
We apply our algorithm to searching decoder architectures in semantic segmentation and gain a set of DF-Seg networks.
Figure~\ref{fig:segmentation_comparison} shows a comparison of our DF-Seg networks and other methods. It can be seen that
our segmentation networks achieve new state-of-the-art in real-time urban scene parsing tasks.

To sum up, we make following contributions to network architecture search:
\begin{itemize}
	\vspace{-2mm}
	\setlength\itemsep{0em}
	\item We are among the first to investigate the problem of balancing speed and accuracy of network architectures for network architecture search. By pruning the search space with a partial order assumption, our ``Partial Order Pruning'' algorithm can efficiently lift the boundary of speed/accuracy trade-off.
	
	\item We present several DF networks that provide both high accuracy and fast inference speed on {target embedded device TX2.}
    The accuracy of our DF1/DF2A networks exceeds ResNet18/50 on ImageNet validation set, but the inference latency is  $43\%$ and $39\%$ lower, respectively.
 
	\item We apply the proposed algorithm to searching decoder architectures for a segmentation network. Together with DF backbone networks, we achieve new state-of-the-art in real-time segmentation on both high-end GPUs and {target embedded device TX2}. On GTX 1080Ti, our DF1-Seg network achieves $106.4$ FPS at resolution $1024\times 2048$ with $\text{mIoU}_\text{class}$ $74.1\%$. On TX2, our DF1-Seg network achieves $21.8$ FPS at resolution $1280\times 720$, \emph{i.e.} 720p. 
\end{itemize}

\vspace{-5mm}
\section{Related Work}
\paragraph{Efficient Network Design}
Group convolution plays a key role in current efficient CNN architecture design~\cite{ma2018shufflenet,howard2017mobilenets,sandler2018inverted}.
MobileNet\_V2~\cite{sandler2018inverted} adopts an inverted residual module  that uses group convolutions to reduce the FLOPs during inference. 
ShuffleNet~\cite{zhang2018shufflenet} uses pointwise group convolution and channel shuffle operation to reduce FLOPs while maintaining accuracy.
\cite{ma2018shufflenet} points out that there is a discrepancy between indirect metric (FLOPs) and direct metric (inference speed), and proposes four guidelines for efficient network design. 
These works design a single architecture without considering the target platform while
our algorithm explicitly takes platform characteristics into consideration.

\begin{figure*}[!t]
	\vspace{-5mm}
	\centering
	\subfigure[]{
		\includegraphics[width=0.66\linewidth]{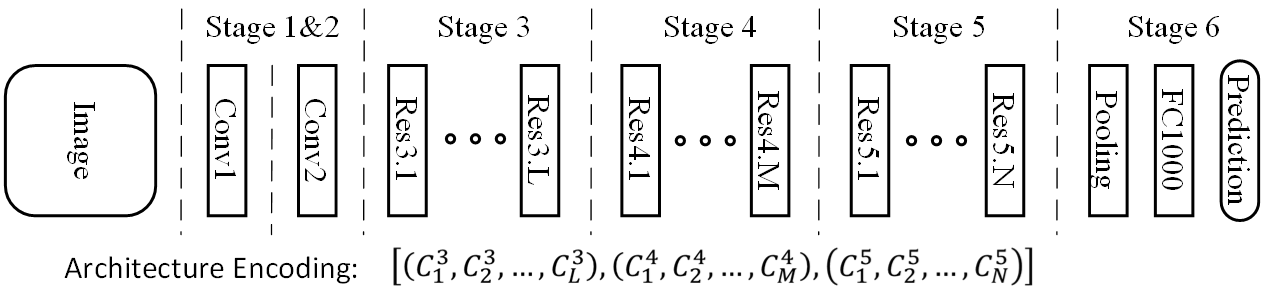}
		\label{fig:search_space}
	}
	\quad \ 
\subfigure[]{
	\includegraphics[width=0.28\linewidth]{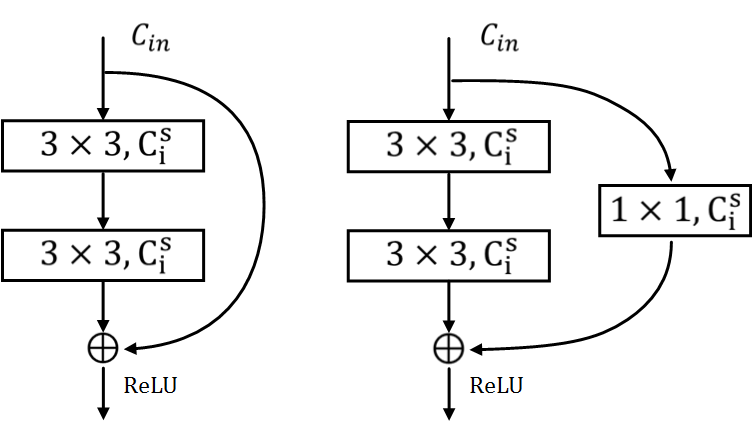}
	\label{fig:basic_block}
}
	\vspace{-2mm}
	\caption{\small (a) General network architecture. (b) The residual block used throughout this paper.  }
	
	\vspace{-6mm}
\end{figure*}

\vspace{-5mm}
\paragraph{Neural Architecture Search}
Automatic network architecture search is often tackled with either reinforcement learning~\cite{zoph2018learning,zoph2017neural} or evolutionary algorithms~\cite{real2018regularized,real2017large}. 
They require huge computational resources, and the obtained networks are relatively slower than manually designed ones~\cite{ma2018shufflenet,dong2018dpp}, even with comparable FLOPs. 
{More recently, several gradient-based algorithms~\cite{darts,fbnet,cai2018proxylessnas,dong2019search} are proposed to reduce the architecture search cost.
Besides, a few works~\cite{dong2018dpp,fbnet,cai2018proxylessnas} also take platform-related objectives into consideration in architecture search. }
Although their goal is somewhat similar to ours, our work differs in that we pursue the balance of width and depth of a network, instead of searching the architecture in each individual block.

\vspace{-5mm}
\paragraph{Real-time Semantic Segmentation}
Most semantic segmentation methods~\cite{zhao2017pyramid,chen2018searching,chen2018encoder} aim at high performance but with relatively slow inference speed. 
For fast semantic segmentation, early works~\cite{paszke2016enet,badrinarayanan2017segnet} employ relatively shallower backbone networks and lower image resolution, offering fast inference speed but poorer accuracy. 
More recently, ICNet~\cite{zhao2018icnet} uses the image cascade to speed up inference, in which 
pre-trained deep CNNs are only applied to the images with lowest resolution. 
BiSeNet~\cite{yu2018bisenet} employs a context path to obtain a sufficient receptive field, and an additional spatial path with a small stride to preserve spatial information. 
None of them attempts to accelerate inference by improving the backbone network, or considers the characteristics of target platforms. Comparatively, our algorithm explicitly takes platform characteristics into consideration, and aims at better speed/accuracy trade-off in both backbone network and decoder network.

\vspace{-5mm}
\paragraph{Model Acceleration}
Some researchers try to accelerate inference of a pre-trained network via quantization~\cite{wu2016quantized}, pruning~\cite{guo2016dynamic}, factorization~\cite{liu2015sparse}, \textit{etc}. For example, NetAdapt~\cite{yang2018netadapt} automatically adapts a pre-trained CNN to a mobile platform given a resource budget.
Compared with them,  we try to balance the width and depth of the overall architecture.

\vspace{-3mm}
\section{Partial Order Pruning}
\subsection{Search Space}
\label{subsecoverall}
We provide a general network architecture in our search space, as shown in Figure~\ref{fig:search_space}. 
It consists of 6 stages to perform classification from input images. 
Stages 1$\sim$5 down-sample the spatial resolution of the input tensor with a stride of 2, and stage 6 produces the final prediction with a global average pooling and a fully connected layer. 
Stages $1\&2$  extract common low-level features on large tensor size, which brings heavy computation burden. In pursuit of an efficient network, we only use one convolution layer in stage 1\&2, \emph{i.e.} $Conv1$ and $Conv2$. We empirically find this is enough for achieving good accuracy.
For stages 3, 4, 5, each consists of L, M, N residual blocks, where L, M ,N are integers, \emph{i.e.} $L,M,N \in \mathbb{N}$. Different settings of L/M/N lead to different network depths.
The width (number of channels) of the $i$-th residual block in stage $s$ is denoted as $C^{s}_{i}$.
Therefore, an architecture can be encoded as shown in Figure~\ref{fig:search_space}. 
In practice, we restrict $C^{s}_{i} \in\{64,128,256,512,1024\}$.
We empirically restrict the width of a block to be no narrower than its preceding blocks.
Throughout this paper, we use the basic residual block proposed in~\cite{he2016deep} if not mentioned otherwise. As shown in Figure~\ref{fig:basic_block}, the building block consists of two convolution layers and a shortcut connection. 
An additional projection layer is added if the size of input does not match the output tensor. All convolutional layers are followed with a batch normalization~\cite{ioffe2015batch} layer and ReLU nonlinearity.

\begin{figure}[!t]
	\centering
	\label{fig:preliminary}	
	\subfigure[]{
		\includegraphics[width=0.45\linewidth]{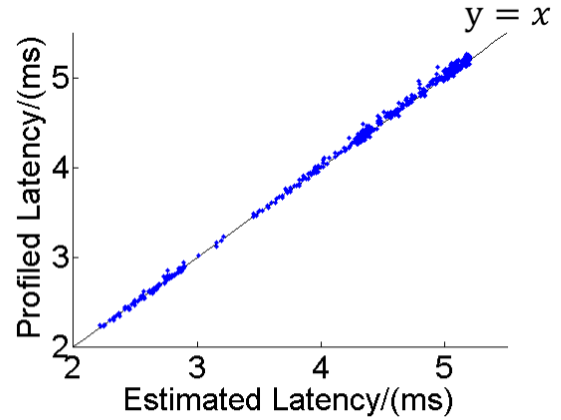}
		\label{fig:compare_tx2time}
	}
	\subfigure[]{
	\includegraphics[width=0.45\linewidth]{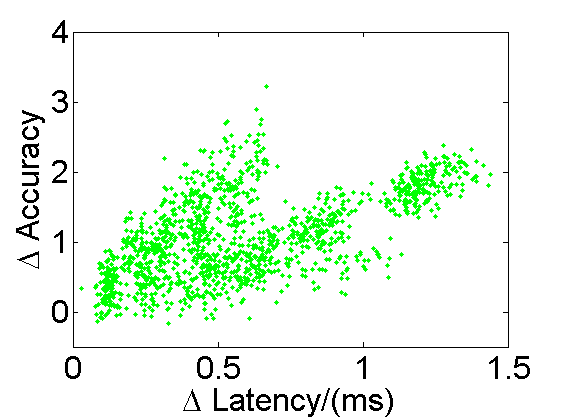}
	\label{fig:validation_partial_order}
	}
	\vspace{-2mm}
	\caption{\small Preliminary experiments. (a) Our latency estimation is highly close to the actual profiled latency. (b) We experimentally verified that partial order assumption is generally true for efficient architectures of our concern.}
	\vspace{-4mm}
\end{figure}


\vspace{-2mm}
\subsection{Latency Estimation}
\label{sec:subspace}
The set of all possible architectures, with different depths (number of blocks) and widths (number of channels per block), is denoted as $\mathbb{S}$ and usually referred to as the search space in neural architecture search~\cite{darts,zoph2018learning}. The latency of architectures in $\mathbb{S}$ can vary from very small to positive infinity.
But we only care about architectures in a subspace $\widehat{\mathbb{S}} \subset \mathbb{S}$, which provide latency in the range $[T_{min},T_{max}]$.

We employ the profiler provided by TensorRT library to obtain layer-wise latency of a network.
We empirically find that a block with a specific configuration (\emph{i.e.} input/output tensor size) always consumes the same latency.
Thus we can construct a look-up table $Latency\left (c_i,h_i,w_i,c_o,h_o,w_o\right)$ providing latency of each block configuration, where $c_i/c_o$ is the number of channels in input/output tensor, and $h_i/w_i/h_o/w_o$ is the corresponding spatial size. 
For example, $Latency\left (32,112,112,64,56,56\right )=0.143$ms on TX2. 
By simply summing up the latency of all blocks, we can efficiently estimate the latency $Lat(x)$ of an architecture $x \in \mathbb{S}$. 
In Figure~\ref{fig:compare_tx2time}, we compare the estimated latency with the profiled latency. 
It shows our latency estimation is highly close to the actual profiled latency. 
All architectures with latency ranging $[T_{min},T_{max}]$ form the subspace $\widehat{\mathbb{S}}$.
This subspace construction significantly narrows down our search space, and hence accelerates the architecture selection.

%
%

\subsection{Partial Order Assumption}\label{subsecpoa}
{A partial order is a binary relation defined over a set. It means that for certain pairs of elements $\left (x,y\right )$ in the set, one of the elements $x$ precedes the other $y$ in the ordering, denoted with $x\prec y$.  Here ``partial" indicates that not every pair of elements needs to be comparable.}

\begin{figure}[!t]
	\centering
	\includegraphics[width=0.85\linewidth]{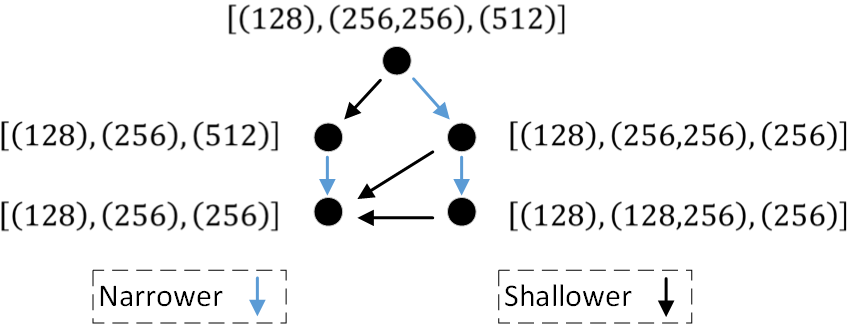}
	\caption{\small {Partial order relations among architectures. An architecture (\emph{e.g.} $[(128),(256),(256)]$) may be narrower than another one with same depth (\emph{e.g.} $[(128),(256),(512)]$), or shallower than another with same width (\emph{e.g.} $[(128),(256,256),(256)]$).}}
	\label{fig:partial_order_relations}
	\vspace{-3mm}
\end{figure}

{We find that there is a partial order relation among architectures in our search space. In Figure~\ref{fig:partial_order_relations}, we follow the architecture encoding in Figure~\ref{fig:search_space}, and illustrate the partial order relation among architectures. 
As explained in Section~\ref{sec:subspace}, $\widehat{\mathbb{S}}$ is a set that contains all architectures in which we are interested.
Let $x,y \in \widehat{\mathbb{S}}$ denote two elements in the set $\widehat{\mathbb{S}}$. If $x$ is shallower than $y$ but they are with the same width, or narrower than $y$ with same depth,
we can borrow the concept from the order theory, and say that $x$ precedes $y$ in the ordering, denoted as $x\prec y$. 
In the rest of this paper, we also call $x$ a \emph{precedent} of $y$ if $x\prec y$.} 
Let $Acc(x)$ and $Lat(x)$ denote the accuracy and latency of the architecture $x$. 
Then the partial order assumption of architectures can be summarized as 
\begin{equation}
\vspace{-2mm}
\label{eqn:assumption}
  Lat(x)\le Lat(y),Acc(x) \le Acc(y), 
\end{equation}
where $\forall x,y \in \widehat{\mathbb{S}},x\prec y$.
Formula (1) assumes that the latency and accuracy of an architecture are both higher than those of its precedents.
This assumption may not hold for very deep networks that contain hundreds of layers~\cite{he2016deep}, but it is generally true for the efficient architectures of our concern, \emph{i.e.} $\widehat{\mathbb{S}}$, which is experimentally verified in this work. 
We find all comparable architecture pairs $(x,y), x\prec y$ in our trained architectures (Section~\ref{subsec:backbone_results}), and compute the latency difference $\Delta Lat = Lat(y) - Lat(x)$ and accuracy difference $\Delta Acc = Acc(y) - Acc(x)$ in each pair.  
As  shown in Figure~\ref{fig:validation_partial_order}, most points locate in the first quartile. 
This means the accuracy of the precedent $x$ is lower, for almost all comparable pairs.
We also notice that a few points locate in the second quartile, but the lower limit of $\Delta Acc.$ is $-0.1\%$, which is negligible considering the randomness during training. 
The above experimental results validate the reasonableness of our partial order assumption.
This assumption can be utilized to prune the architecture search space, and speed up the search process significantly.

\subsection{Partial Order Pruning}


Formally, the goal of our architecture searching algorithm is to obtain an architecture with highest accuracy within every small latency range $[T, T+\delta t]$:
\begin{equation}
\vspace{-2mm}
\label{eqn:optimization}
	\max\limits_{x\in \mathbb{{S}}}{Acc(x)}, s.t. Lat(x) \in [T, T+\delta t] 
\end{equation}
where $\delta t$ is a short time period such as 0.1ms. 
Instead of searching at every small latency range, we optimize within the entire latency range $[T_{min}, T_{max}] $.
With our ``Partial Order Pruning''  algorithm, architecture searching at higher latency helps reduce the searching space at lower latency,
and hence speeds up the overall searching process.

\begin{algorithm}[!t]
\footnotesize
\caption{Partial Order Pruning}
\label{alg:pipeline}
\begin{algorithmic}
\STATE {Initialize trained architecture set $D=\varnothing$}
\STATE {Initialize pruned architecture set $P=\varnothing$.}
\REPEAT
\STATE {Random select an architecture $x \in \widehat{\mathbb{S}}\setminus P$.}
\STATE {Train $x$ and obtain its Acc(x).}
\STATE{$D\leftarrow D \cup \{x\}$.}
\FORALL {$w\in D$}
\STATE {$y_w = \arg\min\limits_{y\in D}Lat(y), s.t. Acc(y)\ge Acc(w)$}
\STATE {$\triangle P_w = \{ m\in \widehat{S}|m \prec w,Lat(m) \ge Lat(y_w)\}$}
\STATE {$P\leftarrow P\cup ( \underset{w}{\cup} \triangle P_w)$}
\ENDFOR
\STATE {$B(D) = \{x \in D| \forall m \in D, $\\ $\quad\quad\quad \quad Lat(m) \ge Lat(x) \ or \ Acc(m)\le Acc(x)\} $}
\UNTIL{No change to B(D) for several iterations.}
\end{algorithmic}
\end{algorithm}

\begin{figure}[!t]
	\vspace{-2mm}
	\centering
	\includegraphics[width=0.75\linewidth]{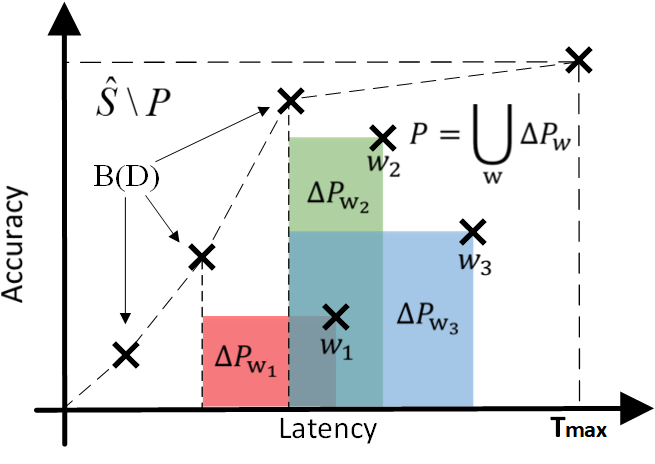}
	\caption{\small We construct pruned search space $P$ with partial order assumption, and prune the search space to be $\widehat{S}\setminus P$. Architectures in $B(D)$ form the boundary for speed/accuracy trade-off we have achieved. (Best viewed in color).}
	\vspace{-2mm}
	\label{fig:update_boundary}
	\vspace{-3mm}
\end{figure}


We use a cutting plane algorithm to optimize the combinational optimization problem in Formula~\eqref{eqn:optimization}.
Algorithm~\ref{alg:pipeline} summarizes the pipeline of our algorithm. 
$D$ is a set containing all trained architectures, and is initialized as empty. 
$P$ denotes the search space pruned from $\widehat{\mathbb{S}}$.
Each time we train a new architecture $x \in \widehat{\mathbb{S}} \setminus P$ and obtain its accuracy $Acc(x)$, we are able to update the pruned search space $P$. 
Figure~\ref{fig:update_boundary} shows how to construct $P$  with the aforementioned partial order assumption.
For each trained architecture $w\in D$, we find the fastest architecture  $y_w \in D$ that provides better accuracy: 
\begin{equation}
 y_w \leftarrow \arg\min\limits_{y\in D}Lat(y), s.t. Acc(y)\ge Acc(w).
\end{equation}
If no $y_w$ is found that satisfies the condition, we continue to process the next $w$.
Let  $\triangle~P_w$ denote the precedents of $w$ with  latency higher than $y_w$, \emph{i.e.} 
\begin{equation}
\triangle P_w=\{m \in \widehat{S}| m \prec w , Lat(m)\ge Lat(y_w)\}.
\end{equation} 
Based on the partial order assumption, a precedent $m$ has lower latency and accuracy, \emph{i.e.} $Acc(m) \le Acc(w)$.
Therefore, even though we do not actually train $m$, we can assume 
\begin{equation}
\forall m \in \triangle P_w, Acc(m)\le Acc(y_w).
\end{equation}
In Figure~\ref{fig:update_boundary}, for all $ m \in \triangle P(w_{i})$,  $i\in \{1,2,3\}$, the $(Lat(m),Acc(m))$ shall locate in the corresponding shadow area.
These architectures in $\triangle P_w$ are very unlikely to provide better speed/accuracy trade-off,
and thus get pruned from the search space to avoid unnecessary training cost.


Given trained architectures $D$, $B(D)$ denotes the architectures that provide best speed/accuracy trade-off in trained architectures:
\begin{eqnarray}
	\vspace{-5mm}
\label{eqn:boundary}
	B(D) &= &\{x \in D| \forall w \in D,  Lat(w) \ge Lat(x)  \nonumber; \\ 
                & &	\ or\ Acc(w)\le Acc(x)\}. 
	\vspace{-5mm}
\end{eqnarray}
Architectures in $B(D)$ form the boundary for speed/accuracy trade-off we can achieve on the target platform.
Figure~\ref{fig:update_boundary} shows $B(D)$ and the corresponding speed/accuracy trade-off boundary.
Intuitively, no architecture in $D\setminus B(D)$ could obtain higher accuracy with lower latency.
By pruning $P$ from the search space $\widehat{\mathbb{S}}$, our algorithm speeds up the architecture search process, 
and lifts the boundary of speed/accuracy trade-off. 
We stop the search process if no change to the $B(D)$ happens for several iterations. 



\subsection{Decoder Design}

\begin{figure}[!t]
\centering
\label{fig:decoder_design}
\vspace{-2mm}
\subfigure[]{
	\includegraphics[width=0.5\linewidth]{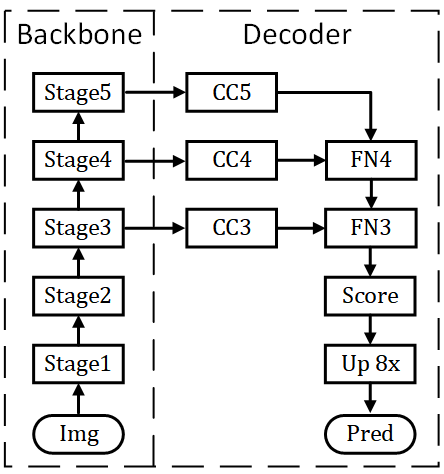}
	\label{fig:decoder_design:a}
}
\subfigure[]{
	\includegraphics[width=0.43\linewidth]{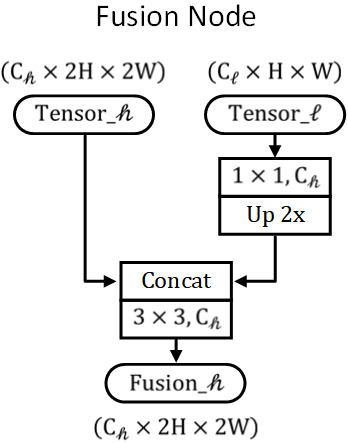}
	\label{fig:decoder_design:b}
}
\caption{\small (a) Overall architecture of segmentation network. (b) Detailed architecture of the fusion node.}
\vspace{-5mm}
\end{figure}

With the proposed Algorithm~\ref{alg:pipeline}, we are able to find backbone architectures that provide best speed/accuracy trade-off on the target platform. 
Given a backbone network, we build semantic segmentation networks as shown in Figure~\ref{fig:decoder_design:a}.
Each stage in the backbone network down-samples the resolution by 2. The resolution of tensors in stage 5 is thus $1/32$ of the input image. We append a pyramid pooling module~\cite{zhao2017pyramid} after the output tensor of stage 5 to improve segmentation performance.
These tensors are then processed by the decoder to produce final prediction. 

We append a $1\times1$ convolution layer after stage 3/4/5 as a ``Channel Controller'' (CC). 
The channel controllers reduce the number of  channels in the corresponding stage without changing its spatial resolution. 
The decoder fuses the tensors in different stages through the fusion nodes. 
The architecture of the fusion node is shown in Figure~\ref{fig:decoder_design:b}. 
A fusion node first projects a low resolution tensor from $C_\ell$ channels to $C_{h}$ channels with a $1\times1$ convolution layer, and then up-samples it by 2. 
We concatenate the up-sampled tensor with a higher resolution tensor, and then process it with a $3\times3$ convolution layer, to fuse the expressive power of different backbone stages.
We fuse the features from stage 3/4/5 and produce a $1/8$ resolution score map.
The score map is then up-sampled by 8 to produce final  per-pixel semantic segmentation prediction.

Let $C^{s}$, $s=3,4,5$ denote the width of each CC. We heuristically set $C \in \{K, 32,64,128,256,512\}$, where $K$ is the number of classes.
Given a backbone network, different settings of channel controllers, \emph{i.e.} $[C^{3},C^{4},C^{5}]$,  lead to different decoder architectures. 
All the possible decoder CC settings form the search space of the decoder architecture. 
Similar to backbone network architectures, we also apply a partial order assumption over the CC settings. 
That is, a narrower decoder is always more efficient and less accurate than a wider one.
Therefore we can also employ the ``Partial Order Pruning" algorithm to lift the speed/accuracy trade-off boundary in the decoder architecture search.

\vspace{-2mm}
\section{Experiment}

\subsection{Experimental Settings}

We adopt two typical kinds of hardware that provide different computational power. 
\begin{itemize}
	\vspace{-2mm}
	\setlength\itemsep{0em}
	\item Embedded device: We use Nvidia Jetson TX2 with an integrated 256-core Pascal GPU as the target embedded device. It provides considerable computational power with limited electrical power consumption.
	\item High-end GPU: We use Nvidia Geforce GTX 1080Ti that provides enormous computing power. We also use GTX Titan X (Maxwell) for fair comparison with previous methods.
	\vspace{-2mm}
\end{itemize}
We adopt two tools to measure inference speed. 
First, we employ the widely used high-performance CNN inference framework TensorRT-3.0.4.
Second, for a fair comparison with ICNet~\cite{zhao2018icnet}, we use the time measure tool Caffe Time, and set the repeating number to $100$ and take the average inference time for comparison.
All experiments are performed under CUDA 9.0 and CUDNN V7.


We conduct experiments on two benchmark datasets. 
The ImageNet~\cite{deng2009imagenet} is a large-scale image classification dataset, which contains over 1.2 million color images in the training set and 50k color images in the validation set. 
The Cityscapes~\cite{cordts2016cityscapes} is a large benchmark dataset for urban scene parsing. It contains $5,000$ images with high quality pixel-level annotations, and is split to $2,975$ for training, $500$ for validation, and $1,525$ for testing. 

\begin{figure}[!t]
	\centering
	\includegraphics[width=0.8\linewidth]{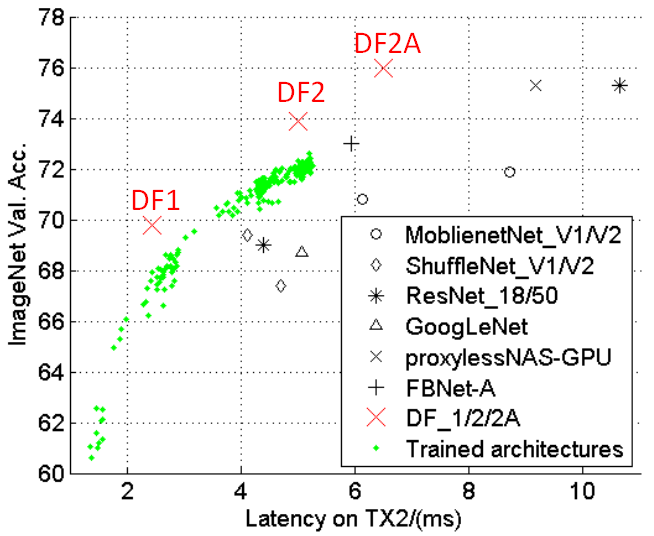}
	\caption{\small Comparison with other popular networks on TX2.}
	\label{fig:backbone_results}
	\vspace{-5mm}
\end{figure}

\subsection{Backbone Architecture Search}
\label{subsec:backbone_results}
In contrast to current architecture search algorithms that conduct architecture searching on small datasets, we directly conduct architecture searching on ImageNet.
We use the SGD optimizer with the poly learning rate policy to train models. 
The power is set to $2$, and the momentum is set to $0.9$. 
We use a weight decay of $0.0001$. 
The batch size is set to $2048$. 
We employ random scaling and stretching for data augmentation to relieve overfitting. 
Following~\cite{goyal2017accurate}, we first train each network for $5$ epochs with learning rate $0.1$ as a warm up scheme, and then train for $80$ epochs with an initial learning rate $0.8$.


We conduct backbone architecture searching experiments on TX2 platform.
During searching, we evaluate the single crop Top-1 accuracy on ImageNet validation set and the inference latency at resolution $224\times224$.
We are interested in the efficient architectures with latency falling in the range $[1ms,5ms]$, and construct the search space $\widehat{\mathbb{S}}$ accordingly (Section~\ref{sec:subspace}).
We conduct architecture search with Algorithm~\ref{alg:pipeline}, and stop the search process when no remarkable boundary update is found during the search. 
The resulting speed/accuracy trade-off boundary is considered to be nearly optimal in our search space $\widehat{\mathbb{S}}$ on the target platform TX2.
We train $\sim200$ networks in total, as shown in Figure~\ref{fig:backbone_results}.  
With the training configuration kept unchanged during architecture search, we train two representative network architectures with additional supervision~\cite{szegedy2015going} and more epochs, to further improve their accuracy. 
The resulting models are referred to as DF1, DF2. 
We further replace some of the building blocks in DF2 from basic block in Figure~\ref{fig:basic_block} to bottleneck block~\cite{he2016deep}. The resulting network is denoted as DF2A. 
Figure~\ref{fig:backbone_results} and Table~\ref{tab:comparison_df} give a comparison of our DF networks and popular models\footnote{\label{shuffle}We report latency with our re-implementation.} on the target platform TX2.
Table~\ref{tab:arch_df} shows detailed architectures of these three DF networks.
Training with more sophisticated methods, \emph{e.g.} dropout or label smoothing, may produce higher accuracy, which however is not the focus of this paper.

\begin{table}[!t]
\centering
\footnotesize
\renewcommand\arraystretch{1.1}
\begin{tabular}{l|cccc}
\toprule
Model         & Top1 Acc. &Latency (ms) & FLOPs \\ \midrule
ShuffleNet\_V2~\cite{ma2018shufflenet}          &  69.4\%         & 4.1        & 146M    \\
ResNet-18~\cite{he2016deep}                          & 69.0\%           & 4.4       &   1.8G    \\
ShuffleNet\_V1~\cite{zhang2018shufflenet}     & 67.4\%           & 4.7       & \bf{140M}  \\
GoogLeNet~\cite{szegedy2015going}             & 68.7\%           & 5.1        &  1.43G     \\
MobileNet\_V1~\cite{howard2017mobilenets} & 70.8\%           & 6.1        &     569M  \\
MobileNet\_V2~\cite{sandler2018inverted}      & 71.9\%           & 8.7        &      300M \\
ResNet-50~\cite{he2016deep}                           & 75.3\%          & 10.6     &  3.8G     \\ \midrule
FBNet-A~\cite{fbnet}      &   73.0\%   &   5.9 & 249M \\
ProxylessNAS-GPU~\cite{cai2018proxylessnas}  &  75.1\%    &   9.3    &   -    \\
NASNet-A~\cite{zoph2018learning}    &   74.0\%  &    20.7     &   564M \\   
PNASNET-5~\cite{liu2018progressive}   &   74.2\%  &    27.6     &   588M \\ \midrule
DF1           & 69.8\%           & \bf{2.5}          &   746M    \\
DF2           & 73.9\%           & 5.0            &   1.77G   \\
DF2A          & \bf{76.0\%}           & 6.5          &   1.97G    \\ \bottomrule
\end{tabular}
\vspace{1mm}
\caption{\small Comparison with other popular networks on TX2.}
\label{tab:comparison_df}
\vspace{-5mm}
\end{table}

\begin{table*}[!t]
\vspace{-5mm}
\centering
\footnotesize
\renewcommand\arraystretch{1.1}
\begin{tabular}{m{1cm}<{\centering}|m{1cm}|m{1.5cm}<{\centering}|m{2.3cm}<{\centering}|m{2.3cm}<{\centering}|m{2.3cm}<{\centering}}
\toprule
Stage & Layer & Output size      &  DF1 & DF2 &DF2A \\ \midrule 
1 & Conv1       & $112\times 112$ & $3\times 3,32$ & $3\times 3,32$ & $3\times 3,32$ \\ 
2 & Conv2       & $56\times 56$     & $3\times 3,64$ & $3\times 3,64$ & $3\times 3,64$ \\ 
3 & Res3\_x    & $28\times 28$     & $\begin{bmatrix} 3\times  3,64\\ 3\times  3,64 \end{bmatrix} \times 3$                                                                                                                             & $\begin{bmatrix} 3\times  3,64\\ 3\times  3,64 \end{bmatrix} \times 2$ $\begin{bmatrix} 3\times  3,128\\ 3\times  3,128 \end{bmatrix} \times 1 $ & $\begin{bmatrix} 3\times  3,64\\ 3\times  3,64 \end{bmatrix} \times 2$ $\begin{bmatrix} 3\times  3,128\\ 3\times  3,128 \end{bmatrix} \times 1 $                                                                 \\ 
4 & Res4\_x    & $14\times 14$     & $\begin{bmatrix} 3\times  3,128\\ 3\times  3,128 \end{bmatrix} \times 3$ & $\begin{bmatrix} 3\times  3,128\\ 3\times  3,128 \end{bmatrix} \times 10$ $\begin{bmatrix} 3\times  3,256\\ 3\times  3,256 \end{bmatrix} \times 1$  & $\begin{bmatrix} 1\times  1,128\\ 3\times  3,128\\ 1\times 1, 512 \end{bmatrix} \times 10$ $\begin{bmatrix} 1\times  1,256\\ 3\times  3,256\\ 1\times 1, 1024 \end{bmatrix} \times 1$\\ 
5 & Res5\_x    & $7\times 7$         & $\begin{bmatrix} 3\times  3,256\\ 3\times  3,256 \end{bmatrix} \times 3$ $\begin{bmatrix} 3\times  3,512\\ 3\times  3,512 \end{bmatrix} \times 1$ & $\begin{bmatrix} 3\times  3,256\\ 3\times  3,256 \end{bmatrix} \times 4$ $\begin{bmatrix} 3\times  3,512\\ 3\times  3,512 \end{bmatrix} \times 2$ & $\begin{bmatrix} 1\times  1,256\\ 3\times  3,256\\ 1\times 1,1024 \end{bmatrix} \times 4$ $\begin{bmatrix} 1\times  1,512\\ 3\times  3,512\\ 1\times 1, 1024 \end{bmatrix} \times 2$ \\ 
6 & FC             & $1\times 1$         & \multicolumn{3}{c}{Global Average Pooling, 1000-d FC, Softmax.}  \\ \midrule 
\multicolumn{3}{c|}{Depth} & $23$ & $43$ & $60$ \\ \bottomrule
\end{tabular}
\vspace{1mm}

\caption{\small Detailed architecture of DF networks.}
\label{tab:arch_df}
\vspace{-5mm}
\end{table*}

Compared with ResNet-18 and GoogLeNet, our DF1 obtains a higher accuracy  $69.8\%$ but the inference latency is $43\%$, $51\%$ lower than two baselines respectively. 
our DF2 has a similar latency  but the accuracy is $4.9\%$ and $5.2\%$ higher than the baselines respectively. 
Furthermore, DF2A achieves a surpassing ResNet-50-level accuracy with a $39\%$ lower latency.
Note we use the same building blocks with ResNet-18/50. 
So we attribute the better speed/accuracy trade-off  to the better balancing between depth and width in our architectures.
Specifically,  our DF1/DF2A are slimmer and deeper than ResNet-18/50 for obtaining the same accuracy. 

MobileNet~\cite{howard2017mobilenets,sandler2018inverted} and ShuffleNet~\cite{zhang2018shufflenet,ma2018shufflenet} are state-of-the-art efficient networks that are designed for mobile applications. 
We also compare our DF networks to them on TX2 in Table~\ref{tab:comparison_df} and Figure~\ref{fig:backbone_results}. 
It can be seen our DF1 achieves higher accuracy but lower inference latency.
The MobileNet/ShuffleNet have less FLOPs but higher latency. 
This is because they have higher memory access cost. The total memory cost (\emph{i.e.} intermediate features) for ShuffleNet\_V2 and DF1 is 4.86$M$ and 2.91$M$ respectively.
This also indicates the FLOPs may be inconsistent with latency on the target platform~\cite{wang2018pelee,ma2018shufflenet}. 
Therefore, taking characteristics of target platform into consideration is necessary for achieving the best speed/accuracy trade-off.

{We also compare our DF networks with other models searched by NAS methods~\cite{zoph2018learning,liu2018progressive,fbnet,cai2018proxylessnas}. As shown in Table~\ref{tab:comparison_df}, NASNet~\cite{zoph2018learning} and PNASNet~\cite{liu2018progressive} have not taken latency into consideration, leading to higher latency. Comparing to FBNet~\cite{fbnet} and ProxylessNAS~\cite{cai2018proxylessnas}, which also take target platform-related objectives into neural architecture search, our DF networks show better speed/accuracy trade-off. This can be explained as (a) DF networks are specifically searched for TX2 platform; (b) FBNet and ProxylessNAS use an inverted bottleneck module, which brings more memory access cost; (c) FBNet and ProxylessNAS aim at searching for better building block architectures while we balance the width and depth of the overall architecture.}

\begin{figure}[!t]
	\centering
	\subfigure[]{
	\includegraphics[width=0.45\linewidth]{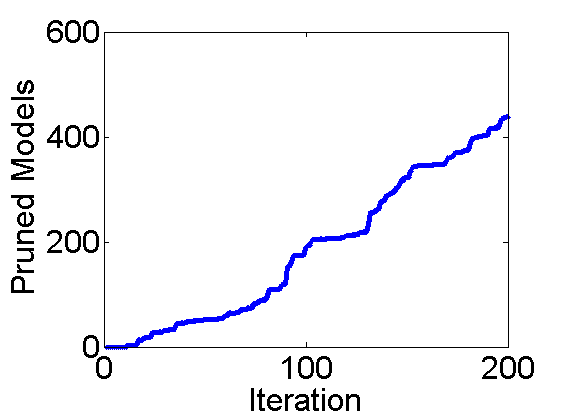}
	\label{fig:pruned_models}
	}
	\subfigure[]{
	\includegraphics[width=0.45\linewidth]{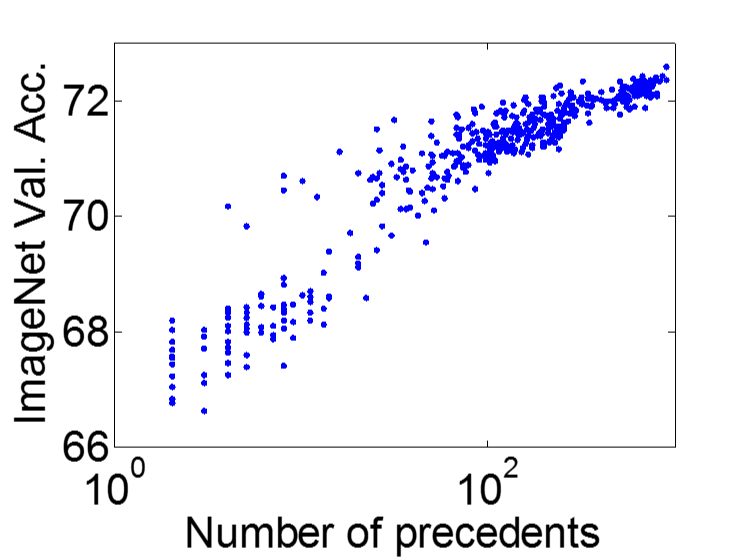}
	\label{fig:accuracy_precedent}
	}
	\vspace{-1mm}

\caption{\small (a) Number of pruned architectures. (b) We empirically find that accuracy of an architecture is correlated to the number of precedents.}
	\vspace{-6mm}
\end{figure}

We then discuss the search efficiency of our proposed algorithm. Figure~\ref{fig:pruned_models} shows the number of pruned architectures in the search process. 
We prune $438$ architectures after training $200$ architectures.
Therefore, our POP algorithm accelerates the architecture search process for $2.2$ times.
Each model takes $5\sim7$ hours on a server with 8-GPUs.
Training $200$ architectures takes $\sim400$ GPU days in total. 
The computational cost of our architecture for searching on ImageNet is lower than the building block architecture searching~\cite{zoph2018learning,real2018regularized} on CIFAR-10 by an order.

Based on our architecture search results, we make following observations.
1) Very quick down-sampling is preferred in early stages to obtain higher efficiency. 
We use 1 convolutional layer in each of stages $1\&2$, and are still able to achieve good accuracy.
2) Down-sampling with the convolutional layer is preferred to the pooling layer for obtaining higher accuracy. 
We only use 1 global average pooling at the end of the network.
3) We empirically find that the accuracy of a network is correlated to the number of its precedents, as shown in Figure~\ref{fig:accuracy_precedent}.
We assume that an architecture with more precedents may have a better balance between depth and width.

\subsection{Decoder Architecture Search}

\begin{figure*}[!t]
\centering


\subfigure[DF1 and TX2]{
	\includegraphics[width=0.23\linewidth]{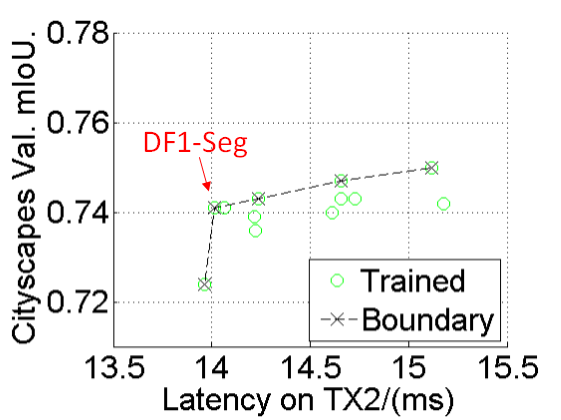}
}
\subfigure[DF1 and GTX 1080Ti]{
	\includegraphics[width=0.23\linewidth]{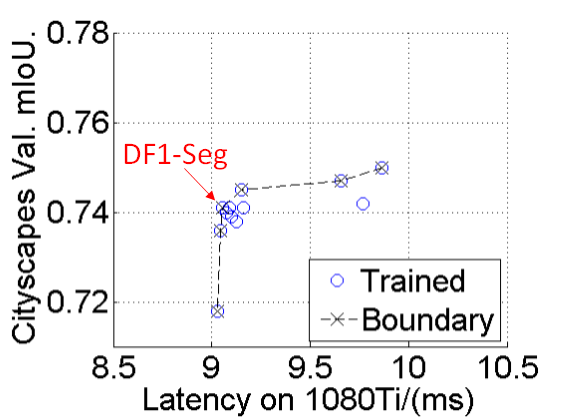}
}
\subfigure[DF2 and TX2]{
	\includegraphics[width=0.23\linewidth]{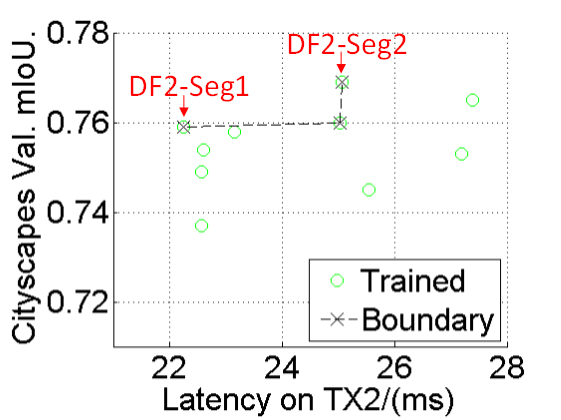}
}
\subfigure[DF2 and GTX 1080Ti]{
	\includegraphics[width=0.23\linewidth]{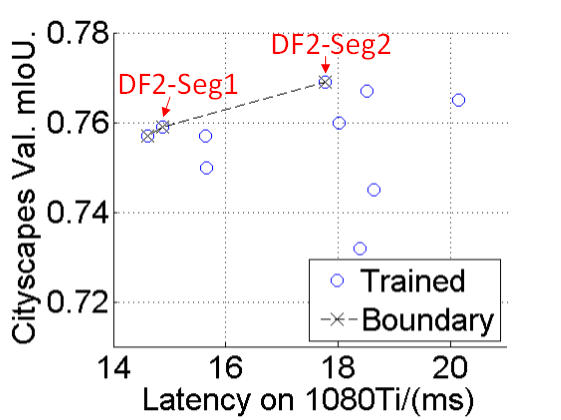}
}
	
	\caption{\small Speed/accuracy trade-off of decoder architecture search results with different backbone networks on different platforms. DF1-Seg/DF2-Seg are two segmentation networks that provide good speed/accuracy trade-off on both TX2 and GTX 1080Ti.}
	\label{fig:decoder_search}
	\vspace{-5mm}
\end{figure*}


With our DF1/DF2 backbone networks, we conduct decoder architecture search experiments on two platforms, 1080Ti and TX2. 
The $\text{mIoU}_\text{class}$  at resolution $1024\times2048$ is taken as a metric of segmentation accuracy. 
The profiler of TensorRT is used to evaluate latency of segmentation networks. 
We evaluate latency at resolution $1024\times 2048$ on 1080Ti, and  $640\times360$ on TX2.

Figure~\ref{fig:decoder_search} shows our decoder architecture search results. 
We select three segmentation networks DF1-Seg, DF2-Seg1, DF2-Seg2 from trained networks that provide good speed/accuracy trade-off on both TX2 and 1080Ti.
The CC setting in the decoder of these three segmentation networks are $[19,32,128]$, $[19,19,32]$, $[19,256,512]$ respectively ($19$ is the number of classes).
Few previous works have reported inference speed on TX2, thus we provide a comparison between our DF-Seg networks and other methods on 1080Ti, as shown in Table~\ref{tab:comparison_dfseg}.
We note~\cite{zhao2018icnet} explicitly explains how they measure inference speed.
Therefore, we add an additional column ``FPS(Caffe)'' in Table~\ref{tab:comparison_dfseg}  for fair comparison. 
Inference speed in the ``FPS(Caffe)'' column is measured by Caffe Time on Titan X (Maxwell) at resolution $1024\times 2048$.

\begin{table}[!t]
\centering
\footnotesize
\renewcommand\arraystretch{1.1}
\begin{tabular}{l|cc|cc}
 \toprule
\multirow{2}{*}{Method} & \multicolumn{2}{c|}{$\text{mIoU}_\text{class}$} & \multirow{2}{*}{FPS} & \multirow{2}{*}{FPS (Caffe)} \\ \cline{2-3} 
                        & val         & test       &                           &                                \\  \midrule
SegNet~\cite{badrinarayanan2017segnet}                  & -           & 56.1       & -                         & -                              \\ 
ENet~\cite{paszke2016enet}                    & -           & 58.3       & -                         & -                              \\ 
ICNet~\cite{zhao2018icnet}                   & 67.7        & 69.5       & -                         & 30.3                           \\ 
ESPNet~\cite{mehta2018espnet}          &      -         &     60.3        &    110 & - \\ 
BiSeNet1$^{\dagger}$~\cite{yu2018bisenet}                & 69.0        & 68.4       & 105.8                     & -                              \\ 
BiSeNet2$^{\dagger}$~\cite{yu2018bisenet}                & 74.8        & 74.7       & 65.5                      & -                              \\ \midrule
DF1-Seg               & 74.1        &       73.0     & 106.4                       & 30.7 \\ 
DF2-Seg1                & 75.9        &    74.8        & 67.2                        & 20.5 \\ 
DF2-Seg2                & \bf{76.9}        &  \bf{75.3}          & 56.3                        & 17.7                            \\  \midrule
DF1-Seg-d8                & 72.4        &       71.4     & \bf{136.9}                       & \bf{40.2}                             \\ \bottomrule
\end{tabular}
\vspace{1mm}
\caption{\small Comparison with other real-time segmentation models on 1080Ti. $^{\dagger}$ means FPS is evaluated at $1536\times 768$.}
\label{tab:comparison_dfseg}
\vspace{-5mm}
\end{table}

Compared with BiSeNet1, our DF1-Seg achieves comparable inference speed, but the $\text{mIoU}_\text{class}$ on val set is $5.1\%$ higher.
Compared with BiSeNet2, DF1-Seg achieves comparable $\text{mIoU}_\text{class}$ on validation set, but the inference speed (FPS) is $1.68$ times faster.
We attribute the better speed/accuracy trade-off of DF1-Seg to its backbone network DF1. 
BiSeNet2 employs ResNet-18 as the backbone network. 
Our DF1 has a comparable accuracy with ResNet-18, but is $1.76$ times faster (2.5ms vs 4.4ms), as shown in Table~\ref{tab:comparison_df}.
Compared with ICNet~\cite{zhao2018icnet}, DF1-Seg achieves comparable inference speed, and the $\text{mIoU}_\text{class}$ is $3.5\%$ higher on test set. 
Our DF2-Seg1 also achieves faster inference speed and better segmentation accuracy than BiSeNet2.
With a wider decoder CC setting ($[19,256,512]$), our DF2-Seg2 achieves the best $\text{mIoU}_\text{class}$ $76.9\%$ on validation set and $75.3\%$ on test set at $56.3$ FPS.

We obtain an even faster segmentation network by dropping the final up-sampling layer, and produce a prediction at $1/8$ of input resolution.
The images to segment are then up-sampled by $8$ times with nearest neighbor interpolation, which can be implemented very efficiently.
We then obtain a DF1-Seg-d8 network that achieves $136.9$ FPS on 1080Ti. The $\text{mIoU}_\text{class}$ on test set ($71.4\%$) is still $1.9\%$ and $3\%$ better than ICNet ($69.5\%$) and BiSeNet1 ($68.4\%$) respectively. 

For fair comparison with previous methods, 
we compare inference speed on Titan X (Maxwell) at different resolution, as shown in Table~\ref{tab:comparison_dfseg_titanx}. Our DF1-Seg and DF1-Seg-d8 achieve $59.9$ FPS and $75.9$ FPS at resolution $1920\times 1080$, \emph{i.e.} 1080p. 
Based on the above experimental results, the DF-Seg networks achieve new state-of-the-art in real-time segmentation on high-end GPU, demonstrating  better speed/accuracy trade-off is achieved.



Previous works~\cite{badrinarayanan2017segnet,paszke2016enet} mostly adopt TX1 to analyze their inference speed. 
In Table~\ref{tab:comparison_dfseg_tx2}, we provide a detailed inference speed analysis on TX2. 
Our DF1-Seg/DF1-Seg-d8 achieve $21.8$ FPS and $29.9$ FPS at resolution $1280\times 720$, \emph{i.e.} 720p.

\begin{table}[!t]
\centering
\footnotesize
\renewcommand\arraystretch{1.1}
\begin{tabular}{m{2cm}|m{1.5cm}<{\centering}m{1.5cm}<{\centering}m{1.5cm}<{\centering}}
\toprule
    \multicolumn{1}{c|}{Method}                 & \tabincell{c}{$640 \times 360$ \\ ms\ /\ FPS} & \tabincell{c}{$1280 \times 720$ \\ ms\ /\ FPS} & \tabincell{c}{$1920 \times 1080$ \\ ms\ /\ FPS}  \\ \midrule
SegNet~\cite{badrinarayanan2017segnet}            &      $69/14.6$       &     $289/3.5$        &     $637/1.6$     \\ 
ENet~\cite{paszke2016enet}               &      $7/135.4$       &      $21/46.8$        &       $46/21.6$             \\ 
BiSeNet-1~\cite{yu2018bisenet}        &            $5/203.5$                 &              $12/82.3$                &              $24/41.4$        \\ 
BiSeNet-2~\cite{yu2018bisenet}        &            $8/129.4$                 &               $21/47.9$               &              $43/23$        \\ \midrule
DF1-Seg        &           $3.65/274.0$                  &              $8.24/121.4$                 &               $16.70/59.9$        \\ 
DF2-Seg1        &           $5.88/170.1$                 &             $13.43/74.5$                 &             $27.36/36.5$              \\ 
DF2-Seg2        &           $6.57/152.2$                &            $15.10/66.2$                  &             $31.08/32.2$            \\ \midrule
DF1-Seg-d8                              &              \bf{3.25/307.7}                &             \bf{6.62/151.1}                &              \bf{13.18/75.9}        \\     \bottomrule
\end{tabular}
\vspace{1mm}
\caption{\small Speed analysis on Titan X (Maxwell).}
\label{tab:comparison_dfseg_titanx}
\vspace{-3mm}
\end{table}

\begin{table}[!t]
\centering
\footnotesize
\renewcommand\arraystretch{1.1}
\begin{tabular}{l|ccc}
\toprule
    \multicolumn{1}{c|}{Method}                 & \tabincell{c}{$480 \times 320$ \\ ms\ /\ FPS} & \tabincell{c}{$640 \times 360$ \\ ms\ /\ FPS} & \tabincell{c}{$1280 \times 720$ \\ ms\ /\ FPS}  \\ \midrule
ESPNet~\cite{mehta2018espnet}          &      -/-         &      -/$\sim$20        &    -/- \\ \midrule
DF1-Seg                                    &             9.45/105.8                &             14.01/71.4                &              45.93/21.8      \\ 
DF2-Seg1                                    &             15.32/65.3                &            22.25/44.9                 &            73.32/13.6       \\ 
DF2-Seg2                                    &             16.98/58.9                &            25.07/39.9                 &             82.07/12.2       \\ \midrule
DF1-Seg-d8                              &               \bf{7.48/133.7}                &             \bf{10.79/92.7}                &              \bf{33.41/29.9}        \\     \bottomrule
\end{tabular}
\vspace{1mm}
\caption{\small Speed analysis on TX2.}
\label{tab:comparison_dfseg_tx2}
\vspace{-6mm}
\end{table}


\vspace{-2mm}
\section{Conclusion}
We propose a network architecture search algorithm ``Partial Order Pruning'' , which is able to lift the boundary of speed/accuracy trade-off of searched networks on the target platform. 
By utilizing a partial order assumption, it efficiently prunes the feasible architecture space to speed up the search process. 
We employ the proposed algorithm to search for both the backbone network and decoder network architectures.  
The searched DF backbone newtorks provide state-of-the-art speed/accuracy trade-off on target platforms.
The searched DF-Seg networks achieve state-of-the-art speed/accuracy trade-off on both embedded devices and high-end GPUs.

\section*{Acknowledgement}
Jiashi Feng was partially supported by NUS IDS R-263-000-C67-646,  ECRA R-263-000-C87-133 and MOE Tier-II R-263-000-D17-112.

\newpage
%

{\small
\bibliographystyle{ieee}
\bibliography{egbib}
}

\end{document}